\documentclass[10pt,twocolumn,letterpaper]{article}

\usepackage{mima}
\usepackage{times}
\usepackage{epsfig}
\usepackage{graphicx}
\usepackage{amsmath}
\usepackage{amssymb}

\usepackage{booktabs}
\usepackage{bm}
\renewcommand{\vec}[1]{\mathbf{#1}}
\newcommand{\mat}[1]{\bm{\mathit{#1}}}
\usepackage{multirow}
\usepackage{threeparttable}

\usepackage{pdfpages}

\usepackage[pagebackref=true,breaklinks=true,letterpaper=true,colorlinks,bookmarks=false]{hyperref}

\mimafinalcopy 

\def\mimaPaperID{4897} 

\ifmimafinal\fi

\begin{document}

\title{$L_2$BN: Enhancing Batch Normalization by Equalizing the $L_2$ Norms of Features}


\author{Zhennan Wang \qquad  Kehan Li \qquad  Runyi Yu \qquad  Yian Zhao  \qquad  Pengchong Qiao \\   
Chang Liu \qquad   Fan Xu \qquad  Xiangyang Ji \qquad  Guoli Song \qquad  Jie Chen\\
}

\maketitle

\begin{abstract}
In this paper, we analyze batch normalization from the perspective of discriminability and find the disadvantages ignored by previous studies: the difference in $l_2$ norms of sample features can hinder batch normalization from obtaining more distinguished inter-class features and more compact intra-class features. 
To address this issue, we propose a simple yet effective method to equalize the $l_2$ norms of sample features. Concretely, we $l_2$-normalize each sample feature before feeding them into batch normalization, and therefore the features are of the same magnitude. Since the proposed method combines the $l_2$ normalization and batch normalization, we name our method $L_2$BN. The $L_2$BN can strengthen the compactness of intra-class features and enlarge the discrepancy of inter-class features. The $L_2$BN is easy to implement and can exert its effect without any additional parameters or hyper-parameters. 
We evaluate the effectiveness of $L_2$BN through extensive experiments with various models on image classification and acoustic scene classification tasks. The results demonstrate that the $L_2$BN can boost the generalization ability of various neural network models and achieve considerable performance improvements.
\end{abstract}

\vspace*{-5mm}
\section{Introduction}


Batch Normalization (BN)~\cite{ioffe2015batch} is a milestone in improving deep neural networks. 
Nonetheless, BN has some disadvantages. 
One of them is that BN does not perform well with a small batch size~\cite{wu2018group}.
Another disadvantage is that BN is not suitable for sequence models~\cite{ba2016layer}, such as RNN~\cite{jordan1997serial}, LSTM~\cite{hochreiter1997long}, GRU~\cite{cho2014properties}, and Transformer~\cite{vaswani2017attention}. 
The information leakage~\cite{wu2021rethinking} is also a shortcoming of BN, which means that the models may exploit mini-batch information rather than learn representations that generalize to individual samples~\cite{he2020momentum}. 
From the robustness perspective, BN may increase adversarial vulnerability and decrease adversarial transferability~\cite{benz2021batch}. As these drawbacks have been identified, there have been many approaches to address them to varying degrees~\cite{yan2020towards, ioffe2017batch, ba2016layer, wu2021rethinking}. 

\begin{figure}[t]
  \centering
  \includegraphics[width=\columnwidth]{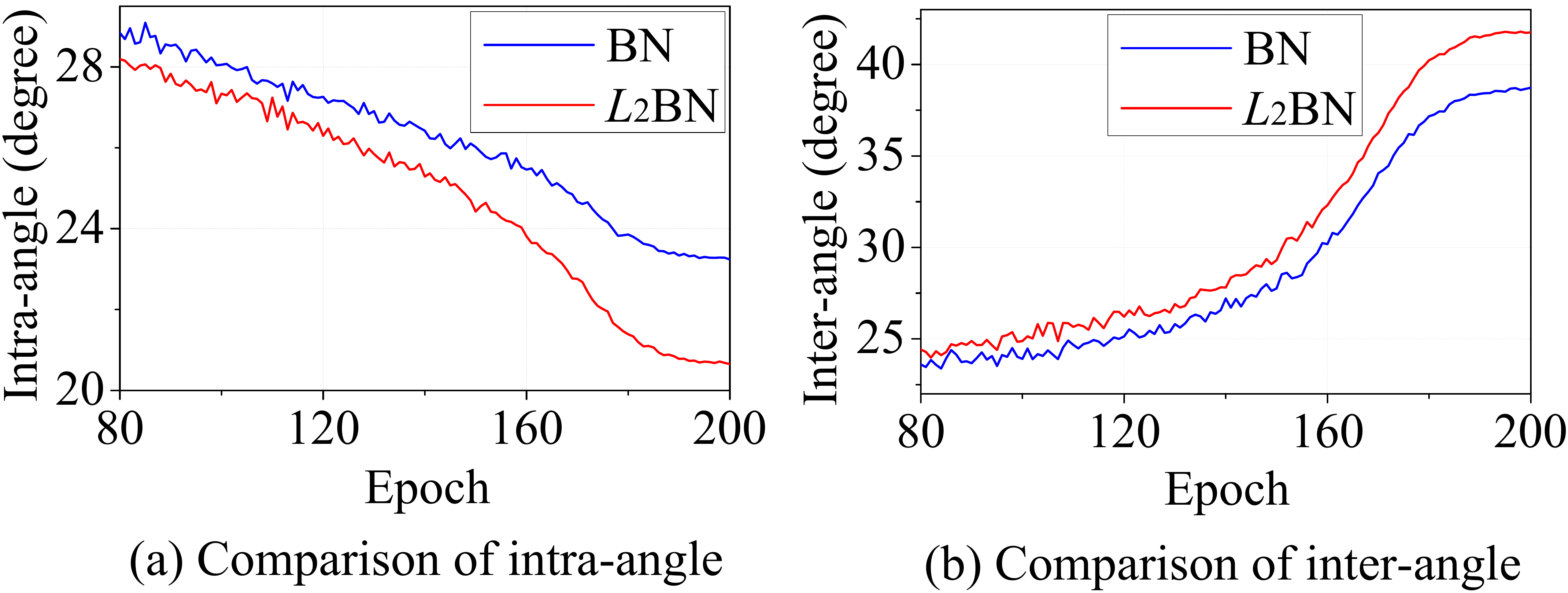}
  \caption{
  (a) The comparison of intra-angle curves indicates that the $L_2$BN can enhance the intra-class compactness. (b) The comparison of inter-angle curves indicates that the $L_2$BN can enlarge the inter-class discrepancy.}
  \label{fig:pull_figure}
  \vspace*{-5mm}
\end{figure}

\begin{figure*}[t]
  \centering
  \includegraphics[width=1.8\columnwidth]{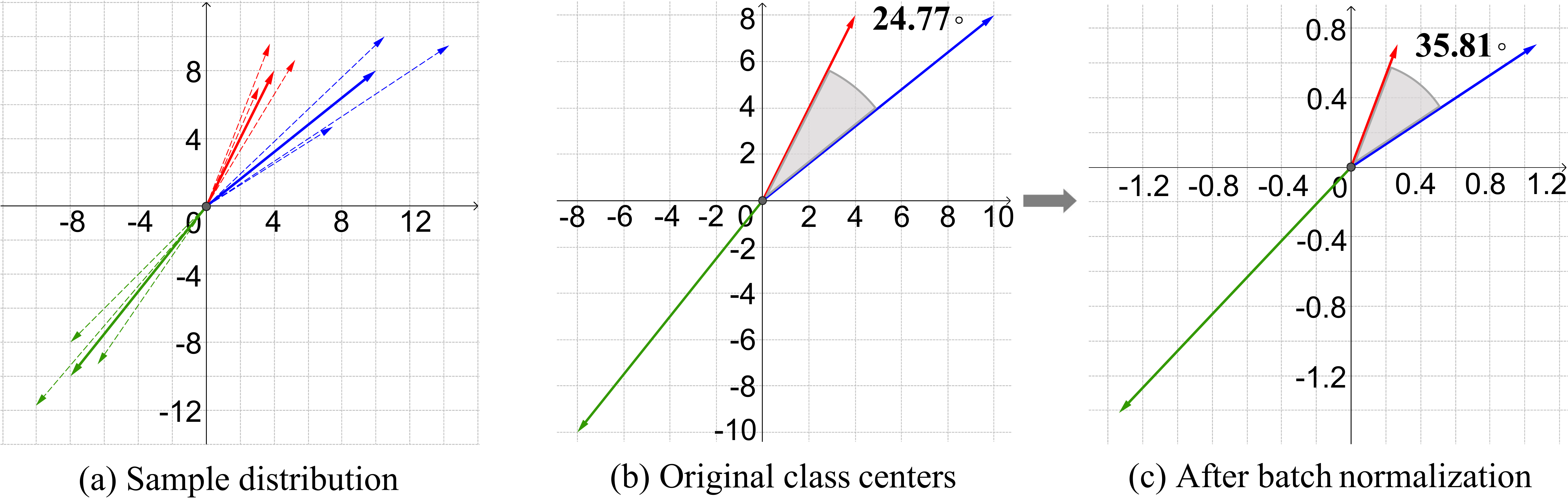}
  \caption{The influence of BN on inter-class features. (a) The dotted vectors with different colors represent features of different classes, and the solid vectors denote the class centers. (b) Since the analysis is on inter-class features, we only consider the class centers for the convenience of analysis.
  The minimum angle between pairwise class centers is 24.77$^{\circ}$. (c) After BN, the minimum angle gets enlarged up to 35.81$^{\circ}$ and will not change even with more identity mapping layers, indicating that BN can not maximize the discrepancy of inter-class features.}
  \label{fig:inter_bn}
\end{figure*}

In this paper, we find that BN has other overlooked shortcomings. 
First, BN does not enlarge the discrepancy of inter-class features to the maximum possible extent. 
Taking Figure~\ref{fig:inter_bn} (b) and (c) as an example,  after the BN, the minimum angle between pairwise class centers increases from 24.77$^\circ$ to 35.81$^\circ$. However, the minimum angle can reach 120$^\circ$ theoretically. Second, BN makes the intra-class features less compact. 
Taking Figure~\ref{fig:intra_l2bn} (a) as an example, 
the intra-class features are similar in orientation originally.  
After the transformation of BN, they are separated in direction. 
Through the forward pass of multiple layers, this intra-class separation may be magnified. 
To the best of our knowledge, we are the first to discover the disadvantages of BN from the perspective of 
discriminability. 

To address these issues of BN, we propose a simple and intuitive approach.  
As shown in Figure~\ref{fig:L2BN_figure}, we just make the $l_2$ norms of features identical before feeding them into BN. 
Since our method combines the $l_2$ normalization and batch normalization, we call it $L_2$BN.  
There are several advantages of $L_2$BN: 
(a) It can continuously broaden the minimum angle between pairwise class centers, as shown in Figure~\ref{fig:inter_l2bn}. Therefore, the $L_2$BN can enlarge the discrepancy of inter-class features. 
(b) The $L_2$BN can eliminate the intra-class separation caused by the difference in $l_2$ norms of sample features, as shown in Figure~\ref{fig:intra_l2bn} (b). 
(c) The $L_2$BN is easy to implement without any extra parameters and hyper-parameters.

To verify the effect of $L_2$BN, we adopt the measures of intra-class compactness and inter-class discrepancy used in ArcFace~\cite{deng2019arcface}. 
The intra-class compactness is measured by intra-angle, which is defined as the mean of angles across features w.r.t. their respective class feature centers. 
The inter-class discrepancy is measured by inter-angle, which is defined as the mean of minimum angles between each class feature center and the others. 
We plot the intra-angle and the inter-angle of training data in Figure~\ref{fig:pull_figure} (a) and (b) respectively, taking the ResNet-110~\cite{he2016deep} model trained on CIFAR100~\cite{krizhevsky2009learning} as an example. 
During the whole training process, the $L_2$BN model achieves a smaller intra-angle than the BN model consistently, indicating that the $L_2$BN obtains more compact intra-class features. 
Furthermore, the $L_2$BN model gradually gets a larger inter-angle than the BN model, indicating that the $L_2$BN obtains more distinguished inter-class features. 
Overall, the $L_2$BN is able to enhance the intra-class compactness and inter-class discrepancy simultaneously, and therefore the feature discrimination and generalization capability are strengthened. 

In practice, the implementation of $L_2$BN is very simple and requires only a few lines of code. 
To exhibit the effectiveness and generality of $L_2$BN, we conduct extensive experiments with various classical convolutional neural networks on tasks of image classification and acoustic scene classification. 
For both tasks, we replace each BN layer in models with an $L_2$BN layer. 
Experimental results show that the $L_2$BN can generally improve the classification accuracy, decrease the intra-angle, and increase the inter-angle, which demonstrates that the $L_2$BN can enhance the generalizability and the discriminability of neural networks. 
These experiments show that the $L_2$BN is generally useful and can be used as an improved alternative to batch normalization in designing neural networks.

\section{Related Work}

\textbf{Typical Normalization Methods.}  
Since Batch Normalization (BN) was proposed by~\cite{ioffe2015batch}, various normalization methods have emerged. 
Recurrent Batch Normalization~\cite{cooijmans2016recurrent} applies BN to the  hidden-to-hidden transition of recurrent neural networks, improving the generalization ability on various sequential problems. 
Layer Normalization~\cite{ba2016layer} (LN) performs a similar normalization to BN, but on elements across the channel or feature dimension, mainly used in sequential models like plain RNN~\cite{rumelhart1986learning}, LSTM~\cite{hochreiter1997long}, GRU~\cite{cho2014properties}, and Transformer~\cite{vaswani2017attention}. 
Instance Normalization~\cite{ulyanov2016instance} (IN) normalizes activations per channel for individual samples, mainly used in image style transfer~\cite{huang2017arbitrary}. 
Positional Normalization~\cite{li2019positional} (PN) normalizes activations along the channel dimension for generative networks. 
Group Normalization~\cite{wu2018group} (GN) is a middle way between IN and LN. 
GN divides the channels into groups and implements normalization within each group,  mainly used in computer vision~\cite{he2016deep, he2017mask}.

In addition to these, some methods explore combinations of these methods. 
Batch Group Normalization~\cite{zhou2020batch} uses the mixed statistics of GN and BN. Batch-Channel Normalization~\cite{qiao2019micro} wraps BN and GN in a module. 
Divisive Normalization~\cite{ren2016normalizing} proposes a unified view of LN and BN. Switchable Normalization~\cite{luo2018differentiable} is a learning-to-normalize method, which switches between IN, LN, and BN by learning their importance weights in an end-to-end manner. 
To avoid redundant computation, Sparse Switchable Normalization~\cite{shao2019ssn} selects only one normalizer for each normalization layer with the help of SparsestMax, a sparse version of softmax. 
Batch-Instance Normalization~\cite{nam2018batch} learns to adaptively combine BN and IN. 
IBN~\cite{pan2018two} uses IN in some channels or layers and BN in other channels or layers. 
XBNBlock~\cite{huang2022delving} replaces the BN with batch-free normalization, like GN, in the bottleneck block of residual-style networks.

Instead of subtracting the mean and divided by the sample standard deviation, there exist other operations to do normalization. 
$L^1$ batch normalization~\cite{hoffer2018norm} replaces the sample standard deviation with the average absolute deviation from the mean, thus improving numerical stability in low-precision implementations as well as providing computational and memory benefits substantially. 
Like IN, Filter Response Normalization~\cite{singh2020filter} also normalizes the activations of each channel of a single feature map, but only divides by the mean squared norm without subtracting the mean value. 
Similarly, Power Normalization~\cite{shen2020powernorm} divides by the mean squared norm along the batch dimension and is mainly used in Transformer for NLP tasks. 
RMSNorm~\cite{zhang2019root} preserves the re-scaling invariance property of LN but eschews the re-centering invariance property, making it computationally simpler and more efficient than LN. 
ScaleNorm~\cite{nguyen2019transformers} further simplifies RMSNorm by setting only one uniform scale parameter for each layer.

Rather than on activations, some methods normalize weights. 
Weight Normalization~\cite{salimans2016weight} aims at decoupling the magnitudes of those weight vectors from their directions by introducing a specific magnitude parameter for each weight vector. 
Weight Standardization~\cite{qiao2019weight} standardizes the weights in the convolutional layers to accelerate the training, which is motivated by~\cite{santurkar2018does} that shows the smoothing effects of BN on activations.



\noindent\textbf{Improvements of Batch Normalization.} 
Despite so many normalization methods, BN is still the most widely used and generally effective for convolutional neural networks~\cite{he2016deep, huang2017densely}.  
However, the original BN comes with some noticeable disadvantages. 
The most known one is the poor performance when the batch size is relatively small~\cite{wu2018group}, due to the unstable batch statistics. 
To address this issue, Batch Renormalization~\cite{ioffe2017batch} uses population statistics instead of batch statistics while subtly retaining the backpropagation of batch statistics. 
To further stabilize training, Moving Average Batch Normalization~\cite{yan2020towards} substitutes batch statistics by moving average statistics in both forward and backward propagation. 
EvalNorm~\cite{singh2019evalnorm} estimates the corrected normalization statistics during evaluation to address the performance degradation of BN. 
Based on Taylor polynomials, Cross-Iteration Batch Normalization~\cite{yao2021cross} jointly utilizes multiple recent iterations to enhance the estimation quality of batch statistics.  
Our proposed $L_2$BN is also an improvement of BN. 
However, unlike the above methods, $L_2$BN is developed to address the issues of inter-class feature discrepancy and intra-class feature compactness caused by BN, which is ignored by previous work.

\section{Method} \label{method}
In this section, we first define the proposed $L_2$BN. 
Then we analyze the advantages of $L_2$BN. 
Finally, we describe the implementation details of $L_2$BN.  
\subsection{The Proposed $L_2$BN} \label{L2BN}
Assuming the input data is $\mat{X} \in R^{b \times d}$, 
where $b$ denotes the batch size and $d$ denotes the feature dimension of input samples, 
the proposed $L_2$BN is formulated as: 
\begin{equation}\label{equ:l21d}
  \mat{\hat{X}}_i = \frac{\mat{X}_i}{\|\mat{X}_i\|} ,
\end{equation}
\begin{equation}\label{equ:l2bn1d}
  BN(\mat{\hat{X}}_i) = \frac{\vec{\gamma}}{\vec{\delta}} \odot (\mat{\hat{X}}_i - \vec{\mu}) + \vec{\beta} ,
\end{equation}
where $\mat{X}_i \in R^d$ denotes the $i$-th sample feature, 
$\mat{\hat{X}}_i$ denotes the $l_2$-normalized feature vector, $\|*\|$ denotes the Euclidean norm, 
$\vec{\mu} \in R^{d}$  and $\vec{\delta} \in R^{d}$ denote the sample mean and uncorrected sample standard deviation of $l_2$-normalized $\mat{X}$ along the batch dimension respectively, $\gamma \in R^{d}$ and $\beta \in R^{d}$  are learnable affine parameters,  
and $\odot$ is the element-wise multiplication between two vectors. 
That is, we perform $l_2$ normalization for each feature vector before feeding them into BN, and thus the magnitudes of feature vectors become identical. 

\begin{figure}[t]
  \vspace*{-5mm}
  \centering
  \includegraphics[width=\columnwidth]{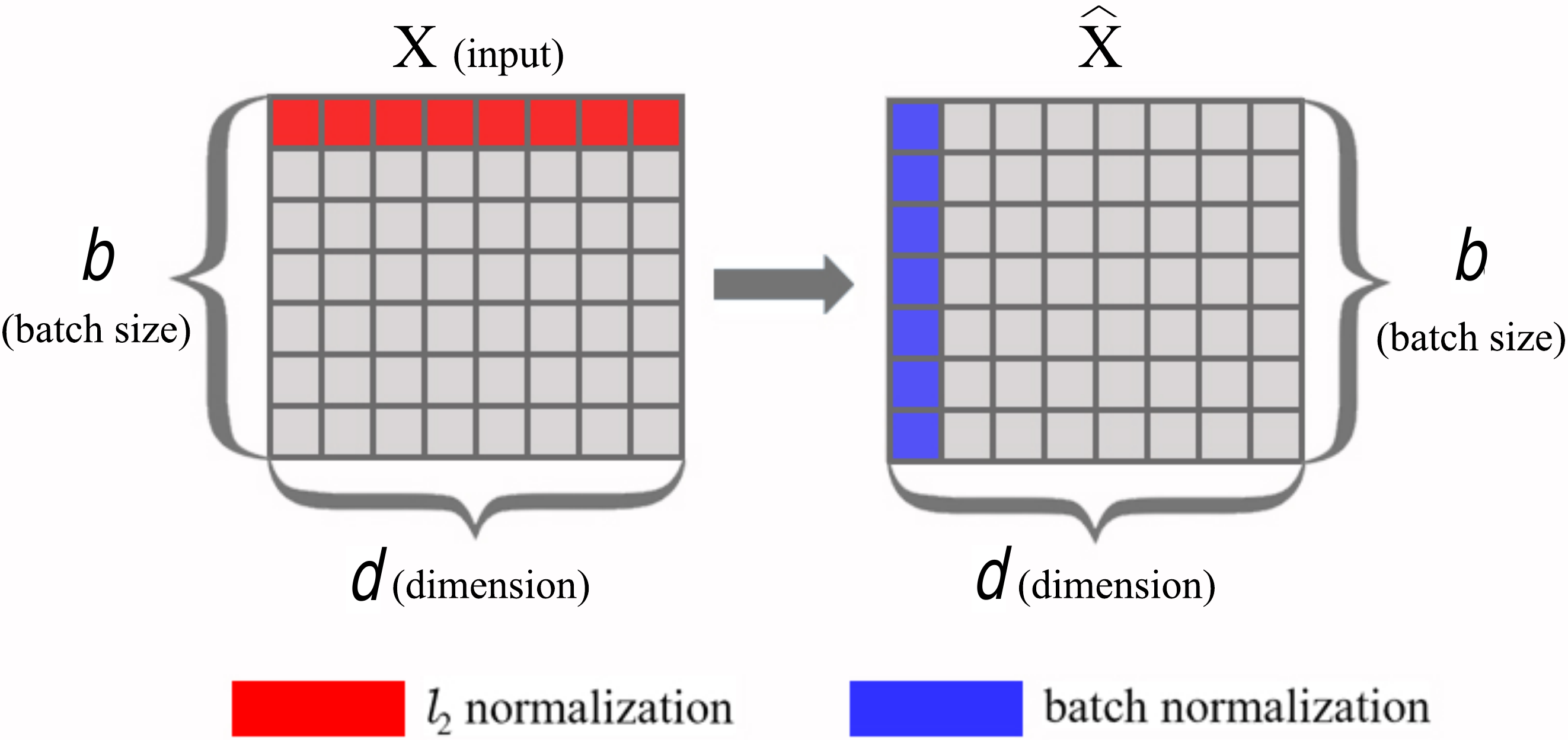}
  \caption{The illustration of $L_2$BN.}
  \label{fig:L2BN_figure}
  \vspace*{-3mm}
\end{figure}

To make the method more intuitive, we visualize the whole process in Figure~\ref{fig:L2BN_figure}. 
Since the proposed method combines the $l_2$ normalization and batch normalization, we name our method $L_2$BN. 
As shown in Figure~\ref{fig:L2BN_figure}, $L_2$BN does not attempt to modify the BN itself, as BN is very helpful 
for the optimization of neural networks~\cite{ioffe2015batch, santurkar2018does}.  
$L_2$BN just implements an additional $l_2$ normalization to make the input samples of BN have equal magnitudes, which leaves the vector orientations as the only difference between samples. 
As analyzed below, this simple but intuitive method can address the issues of inter-class discrepancy and intra-class compactness caused by BN.



\begin{figure*}[t]
  \centering
  \includegraphics[width=1.8\columnwidth]{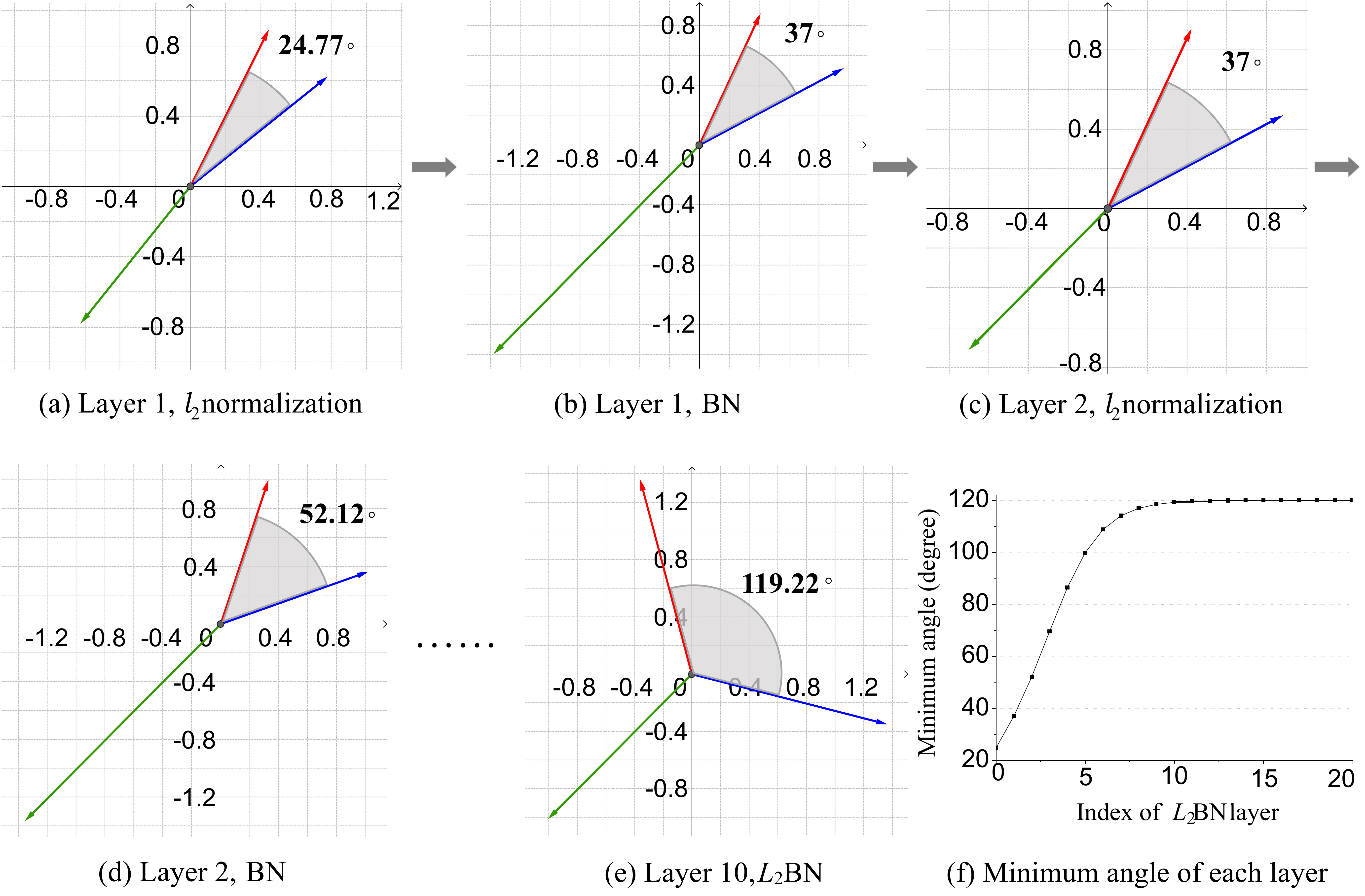}
  \caption{The influence of $L_2$BN on inter-class features. Assume that the layers are identity mappings. 
  (a) After the $l_2$ normalization of $L_2$BN within the first layer, the minimum angle remains the same, but the Euclidean norms become identical. 
  (b) After the batch normalization of $L_2$BN within the first layer, the minimum angle gets enlarged up to 37$^{\circ}$. 
  (c) After the $l_2$ normalization of $L_2$BN within the second layer, the Euclidean norms become the same again. 
  (d) After the batch normalization of $L_2$BN within the second layer, the minimum angle gets enlarged up to 52.12$^{\circ}$. 
  (e) After the $L_2$BN within the tenth layer, the minimum angle gets enlarged up to 119.22$^{\circ}$, indicating that $L_2$BN can enlarge the discrepancy of inter-class features. 
  (f) The minimum angle continues to grow as the number of layers increases until it reaches the maximum.}
  \label{fig:inter_l2bn}
  \vspace*{-3mm}
\end{figure*}

\subsection{Analysis of $L_2$BN} \label{L2BN_analysis}
This section explores the influence of $L_2$BN on the inter-class discrepancy and intra-class compactness. 

\noindent{\textbf{Enlarging the Discrepancy of Inter-class Features.}} 
To facilitate the analysis, we consider the sample distribution in 2-D space, as shown in Figure~\ref{fig:inter_bn} (a). 
Given that this is for analyzing the inter-class discrepancy, we only consider the class center for each class as in Figure~\ref{fig:inter_bn}~(b). 
For simplicity, we assume that the $\gamma$ equals one and $\beta$ equals zero in BN, and each layer in neural network is an identity mapping. 

Because of centering and scaling the elements along the batch dimension~\cite{ioffe2015batch, qi2018face}, the minimum angle between pairwise class centers gets enlarged after the transformation of BN, as shown in Figure~\ref{fig:inter_bn} (c).  
In other words, BN separates the inter-class features to some extent. 
We argue that this is one of the reasons why BN can facilitate the optimization of neural networks. 
However, since the sample mean $\mu$ and sample standard deviation $\delta$  have become 0 and 1 respectively after a single BN operation, the sample distribution will not change even after multiple BN operations. 
Therefore, BN can not separate the inter-class features further, as we can see in Figure~\ref{fig:inter_bn} (c). 

\begin{figure*}[t]
  \centering
  \includegraphics[width=1.9\columnwidth]{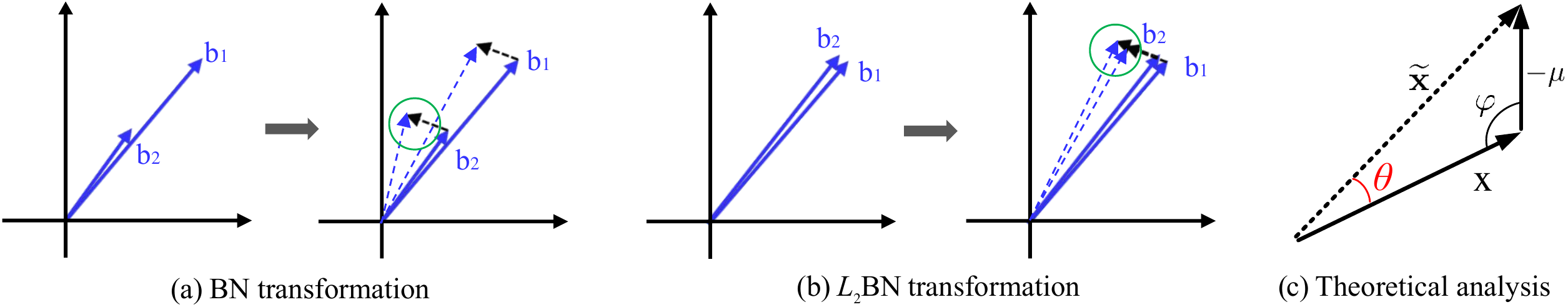}
  \caption{The influence of BN and $L_2$BN on intra-class features. The $b_1$ and $b_2$ vectors belong to the same class, the angle between them is very small but the difference in Euclidean norms is large. The short black dotted vector represents the opposite vector of the shared mean vector $\mu$. 
  For simplicity, we do not consider the influence of sample standard deviation. 
  (a) After BN, the directional difference between $b_1$ and $b_2$ becomes larger, resulting in less compact intra-class features.
  (b) Due to the additional $l_2$ normalization, the directional difference between $b_1$ and $b_2$ is still small after the $L_2$BN transformation. 
  (c) The angle $\theta$ is determined by $\vec{\mu}$, $\varphi$, and the Euclidean length of $\vec{x}$. The $\vec{x}$ denotes the vector before BN and the $\widetilde{\vec{x}}$ denotes the vector after BN. The $\varphi$ denotes the angle between $\vec{x}$ and $\vec{\mu}$.}
  \label{fig:intra_l2bn}
  \vspace*{-3mm}
\end{figure*}

In the case of using $L_2$BN, we visualize the evolution of class centers in Figure~\ref{fig:inter_l2bn}. 
Compared with BN, the advantage of $L_2$BN is that it can continuously expand the minimum angle between pairwise class centers, which benefits from the additional $l_2$ normalization. 
As an illustration, the Euclidean norms of class centers are not identical after the $L_2$BN within the first identity mapping layer in Figure~\ref{fig:inter_l2bn} (a) and (b). 
Thus, the $l_2$ normalization can still change the sample distribution within the second identity mapping layer in Figure~\ref{fig:inter_l2bn} (c).  
In turn, the $l_2$ normalization destroys the distribution state of zero mean and unit variance, which allows the subsequent BN to  remain valid. 
Therefore, the minimum angle between pairwise class centers is further enlarged, as shown in  Figure~\ref{fig:inter_l2bn} (d).   
After several identity mapping layers, $L_2$BN can separate the inter-class features to the maximal extent, as in Figure~\ref{fig:inter_l2bn} (e) and (f). 
As a result, the proposed $L_2$BN can achieve a larger inter-class discrepancy than BN. 

\noindent{\textbf{Enhancing the Compactness of Intra-class Features.}} 
In addition to the impact on inter-class features, the difference in the Euclidean norms of features also affects the compactness of intra-class features after BN. 
For intra-class features, we assume that their orientations are similar and expect them to maintain similar after BN. 
This is a reasonable assumption, since orientations represent  semantic information~\cite{liu2017deep} and intra-class features belong to the same semantic class. 
To illustrate the problem intuitively, we take Figure~\ref{fig:intra_l2bn} (a) as an example, in which the $b_1$ and $b_2$ vectors belong to the same class. 
After the transformation of BN, the intra-class feature vectors with similar orientations but different Euclidean norms are  further separated. 
Through the forward propagation of multiple layers, this separation of intra-class features may be magnified, which leads to less compact intra-class features. 
The proposed $L_2$BN makes the Euclidean norms identical before feeding the features into BN and thus eliminates the impact of difference in Euclidean norms, which is a simple and intuitive approach to address this issue as shown in Figure~\ref{fig:intra_l2bn} (b).  

Besides the above intuitive analysis, we do a brief theoretical analysis. 
The analysis schematic is shown in Figure~\ref{fig:intra_l2bn} (c). 
The $\vec{\mu}$ is calculated per batch and is shared for each sample vector. 
Once the orientation of $\vec{x}$ is given, the angle $\varphi$ between $\vec{x}$ and $\vec{\mu}$ is knowable. 
Also, the $\theta$ is related to the Euclidean length of $\vec{x}$. 
In summary, the $\theta$ can be formulated as $\theta = f(\vec{\mu}, \varphi, \| \vec{x} \|)$. 
Evidently, the variance of $\| \vec{x} \|$ can be passed to the $\theta$. 
The $L_2$BN eliminates the variance of $\| \vec{x} \|$ through an additional $l_2$ normalization. 
Therefore, the variance of $\theta$ is reduced and the compactness of intra-class features is enhanced.

\subsection{Implementation} \label{implementation}
Although the above analysis focuses on the classification layer, the $L_2$BN can also be applied to the hidden layers.  
We experimentally verify that applying $L_2$BN to all layers achieves the greatest accuracy improvement in the Experiments Section~\ref{exp:image}.  
For image tasks like image classification, 2D convolution is commonly used. 
In this case, we perform $l_2$ normalization on the whole feature map of each sample, because  the sample feature is represented by the whole feature map under this setting. 
That is,  Equation~(\ref{equ:l21d}) is replaced with:
\begin{equation}\label{equ:l23d}
\begin{split}
  &\mat{\hat{X}}_i = \frac{\mat{X}_i}{\text{max}(\sqrt{\sum_{j=1}^{C\times H\times W}\mat{X}_{ij}^2}, \epsilon)} , \;\;\;\;\;  \text{or} \\
  &\mat{\hat{X}}_i = \frac{\sqrt{C\times H\times W}*\mat{X}_i}{\text{max}(\sqrt{\sum_{j=1}^{C\times H\times W}\mat{X}_{ij}^2}, \epsilon)} ,
\end{split}
\end{equation}
where $\epsilon$ is a very small number added for division stability. 
The $C$, $H$, and $W$ denote the channel number, height, and width of the feature map respectively. 
Because of the subsequent BN operation, multiplying by $\sqrt{C\times H\times W}$ does not affect the output of $L_2$BN in theory. 
But by doing so, it can prevent floating point underflow in the case of large feature size. 

  \vspace*{-3mm}
\section{Experiments}

\subsection{Image Classification} \label{exp:image}
\noindent\textbf{Experimental Settings.} 
We verify the advantages of $L_2$BN over BN by conducting image classification experiments using convolutional networks. 
We experiment on the CIFAR100 dataset~\cite{krizhevsky2009learning} and the ImageNet-1K  dataset~\cite{russakovsky2015imagenet}. 
On CIFAR100, we employ various classic networks as the backbone models, 
including ResNet-20$\backslash$32$\backslash$44$\backslash$56$\backslash$110~\cite{he2016deep}, 
DenseNet~\cite{huang2017densely} with 40 layers and a growth rate of 12 denoted by DenseNet-40-12, 
DenseNet-BC~\cite{huang2017densely} with 100 layers and a growth rate of 12 denoted by  DenseNet-BC-100-12,
VGG-19~\cite{simonyan2014very} with BN, 
and RegNet-Y~\cite{radosavovic2020designing}. 
On ImageNet-1K, we evaluate our method with ResNet-50, ResNet-101, ResNeXt-50~($32\times 4d$), and ResNeXt-101~($32\times 4d$).


For CIFAR100, we report the mean and standard deviation of the best accuracy over 5 runs with random seeds ranging from 121 to 125, reducing the impacts of random variations. 
For ImageNet-1K, we fix the random seed to 1. 
For a fair comparison, not only the $L_2$BN models but also their BN counterparts are trained from scratch, so our results may be slightly different from the ones presented in the original papers due to different random seeds, software, and hardware settings. 
Other training settings and hyper-parameters are detailed in the supplementary materials.

\begin{table*}[t]
  \centering
  \setlength{\tabcolsep}{3.5mm}
  \begin{tabular}{llcccccc}
    \toprule
     \multirow{2}{*}{Model}  
     &\multirow{2}{*}{\shortstack{BN/\\ $L_2$BN}}  
     &\multirow{2}{*}{Accuracy(\%)$\uparrow$}
     &\multicolumn{3}{c}{Intra and Inter angles (degree)} 
     &\multicolumn{2}{c}{$IIR$}   \\
     \cmidrule(lr){4-6}  \cmidrule(lr){7-8}
     &&&Intra(train)$\downarrow$  &Intra(test)$\downarrow$  &Inter$\uparrow$ &train$\downarrow$ & test$\downarrow$
    \\
    \midrule
    ResNet-20  & BN  & 69.70$\pm$0.30  & 28.15  & 29.04  & 21.70  & 1.297  & 1.338  \\ 
    ResNet-20  & $L_2$BN  & \textbf{70.17}$\pm$0.20  & \textbf{27.61}  & \textbf{28.69}  & \textbf{22.31}  & \textbf{1.238}  & \textbf{1.286}  \\ \hline
    ResNet-32  & BN  & 71.40$\pm$0.20  & 27.43  & 28.75  & 23.65  & 1.160  & 1.216  \\
    ResNet-32  & $L_2$BN  & \textbf{71.75}$\pm$0.21  & \textbf{26.90}  & \textbf{28.39}  & \textbf{24.41}  & \textbf{1.102}  & \textbf{1.163}  \\ \hline
    ResNet-44  & BN  & 72.53$\pm$0.25  & 27.06  & 28.56  & 24.80  & 1.091  & 1.152  \\ 
    ResNet-44  & $L_2$BN  &\textbf{ 73.10}$\pm$0.14  & \textbf{26.28}  & \textbf{28.05}  & \textbf{25.85}  & \textbf{1.016}  & \textbf{1.085}  \\ \hline
    ResNet-56  & BN  & 72.88$\pm$0.25  & 24.76  & 28.91  & 34.09  & 0.726  & 0.848  \\
    ResNet-56  & $L_2$BN  & \textbf{73.92}$\pm$0.20  & \textbf{22.90}  & \textbf{27.73}  & \textbf{35.92}  & \textbf{0.638}  & \textbf{0.772}  \\ \hline
    ResNet-110  & BN  & 74.82$\pm$0.20  & 23.45  & 28.78  & 38.98  & 0.602  & 0.739  \\ 
    ResNet-110  & $L_2$BN  & \textbf{75.32}$\pm$0.44  & \textbf{20.57}  & \textbf{27.00}  & \textbf{41.87}  & \textbf{0.491}  & \textbf{0.645}  \\ \hline
    DenseNet-40-12  & BN  & 75.17$\pm$0.27  & \textbf{27.45}  & \textbf{29.21}  & 23.54  & 1.166  & 1.241  \\ 
    DenseNet-40-12  & $L_2$BN  & \textbf{75.36}$\pm$0.18  & 27.62  & 29.57  & \textbf{24.99}  & \textbf{1.105}  & \textbf{1.183}  \\ \hline
    DenseNet-BC-100-12  & BN  & 77.44$\pm$0.20  & 31.07  & 33.07  & 28.11  & 1.105  & 1.176  \\ 
    DenseNet-BC-100-12  & $L_2$BN  & \textbf{77.87}$\pm$0.10  & \textbf{30.59}  & 33.07  & \textbf{31.28}  & \textbf{0.978}  & \textbf{1.057}  \\ \hline
    VGG-19  & BN  & 73.77$\pm$0.50  & 13.32  & 27.89  & 58.81  & 0.226  & 0.474  \\ 
    VGG-19  & $L_2$BN  & \textbf{73.99}$\pm$0.44  & \textbf{12.05}  & \textbf{27.01}  & \textbf{59.14}  & \textbf{0.204}  & \textbf{0.457}  \\ \hline
    RegNet-Y  & BN  & 78.21$\pm$0.29  & 25.07  & 32.69  & 57.94  & 0.433  & 0.564  \\ 
    RegNet-Y & $L_2$BN & \textbf{78.88}$\pm$0.24 & \textbf{21.39} &\textbf{30.23} & \textbf{59.30} & \textbf{0.361} & \textbf{0.510} \\ 
    \bottomrule
  \end{tabular}
  \caption{Classification results on CIFAR100. We show the accuracy as ``mean$\pm$std''. The $IIR$ denotes the Intra-angle and Inter-angle Ratio, as defined in Equation~(\ref{IIR}).}
  \label{tab:results_cifar100}
\end{table*}

\begin{table}[h]
  \centering
  \begin{threeparttable}
  \begin{tabular}{llc}
    \toprule
    \multicolumn{1}{c}{Model}       &\multicolumn{1}{c}{BN/$L_2$BN}          &Accuracy(\%)  \cr
    \midrule
    ResNet-50       & BN         & 76.74    \\
    ResNet-50       & $L_2$BN    & \textbf{77.32}    \\
    \hline
    ResNet-101      & BN         & 78.43    \\
    ResNet-101      & $L_2$BN    & \textbf{78.81}    \\
    \hline
    ResNeXt-50($32\times 4d$)     & BN             & 77.86         \\
    ResNeXt-50($32\times 4d$)     & $L_2$BN        & \textbf{78.43}         \\
    \hline
    ResNeXt-101($32\times 4d$)    & BN              & 79.14     \\
    ResNeXt-101($32\times 4d$)    & $L_2$BN         & \textbf{79.40}     \\
    \bottomrule
  \end{tabular}
  \end{threeparttable}
  \caption{Classification results on ImageNet-1K.}
    \label{tab:ImageNetAcc}
   \vspace*{-5mm}
\end{table}

\noindent\textbf{Main Results and Analysis.} 
Table~\ref{tab:results_cifar100} and Table~\ref{tab:ImageNetAcc} show the comparison results of the models and their $L_2$BN versions on CIFAR100 and ImageNet-1K, respectively.    

\begin{figure}[t]
  \centering
  \includegraphics[width=\columnwidth]{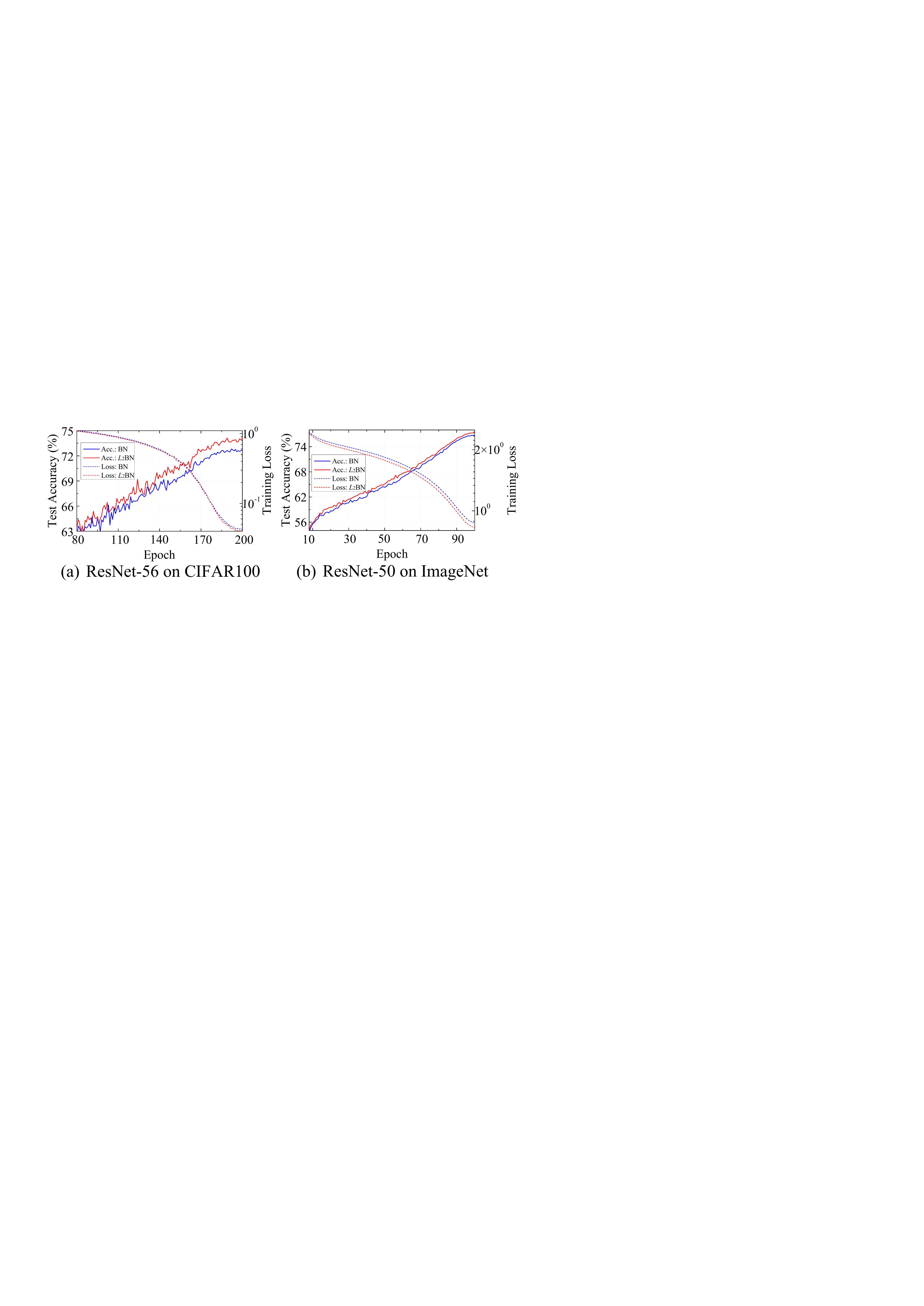}
  \caption{The comparison of training curves. The $L_2$BN is persistently effective.}
  \label{fig:training_curves}
  \vspace*{-4mm}
\end{figure}

In terms of accuracy, the $L_2$BN can improve all the backbone models to varying degrees, regardless of the CIFAR100 or the ImageNet-1K dataset. 
For example, the $L_2$BN can boost the accuracy of ResNet-56 by about 1\% on CIFAR100 and ResNet-50 by about 0.6\% on ImageNet-1K. 
It is worth emphasizing that the $L_2$BN achieves the improvements without any additional parameters or hyper-parameters.  
Due to the pre-attached $l_2$ normalization, the accuracy improvement benefits from the elimination of the difference in $l_2$ norms of sample features. 
To intuitively illustrate the effectiveness of  $L_2$BN, we plot the training curves of ResNet-56 on CIFAR100 and ResNet-50 on ImageNet-1K in Figure~\ref{fig:training_curves}. 
The $L_2$BN can get persistently higher classification accuracy and slightly smaller training loss than the BN baseline.


In Section~\ref{L2BN_analysis}, we analyze that the improvement in accuracy is due to the ability of $L_2$BN to enhance intra-class compactness and inter-class discrepancy. 
In this part, we demonstrate this claim through extensive experiments. 
To measure the intra-class compactness and inter-class discrepancy quantitatively, we adopt the measures used in ArcFace~\cite{deng2019arcface}. 
The intra-class compactness is measured by the mean of angles across features with respect to their respective class feature centers, denoted as intra-angle. 
The inter-class discrepancy is measured by the mean of minimum angles between each class feature center and the other class feature centers, denoted as inter-angle. 
To be more clear, we give the formulations of intra-angle and inter-angle:
\begin{equation}
\vec{c}_i = \frac{1}{N_i} \sum\limits_{j=1}^{N_i} \frac{\vec{x}_{ij}}{\|\vec{x}_{ij}\|} ,
\end{equation}
\begin{equation}
        intra\text{-}angle = \frac{1}{\sum_{i=1}^{C}N_i} \sum\limits_{i=1}^{C} \sum_{j=1}^{N_i} \arccos{(\frac{\vec{x}_{ij}\vec{c}_i^T}{\|\vec{x}_{ij}\| \|\vec{c}_i\|})} ,
\end{equation}
\begin{equation}
        inter\text{-}angle = \frac{1}{C} \sum\limits_{i=1}^{C} \min_{j\in[1,C], j\neq i }
        \arccos{(\frac{\vec{c}_i\vec{c}_j^T}{\|\vec{c}_i\| \|\vec{c}_j\|})} ,
\end{equation}
where $\vec{x}_{ij}\in R^{1\times d}$ denotes the $j$-th sample feature vector of the $i$-th class, 
$N_i$ denotes the number of samples belonging to the $i$-th class, 
$\vec{c}_i \in R^{1\times d}$ denotes the feature center vector of the $i$-th class, and $C$ denotes the number of classes. 

We calculate the intra-angle of training data, the intra-angle of test data, and the inter-angle. 
The inter-angle refers to the one of training data, and we don't calculate the inter-angle of test data. 
We argue that the inter-angle of test data is meaningless because the image classification is a closed-set task. 
For the same reason, we use the class feature centers computed on the training set for the calculation of intra-angle, whether the intra-angle is of training data or test data. 
Note that the intra-angle and inter-angle are only meaningful for the classification layer. 
The results in Table~\ref{tab:results_cifar100} show that the intra-angle is reduced after using the $L_2$BN either on training data or on test data, for most of the models except the DenseNet.  
But for all models, the inter-angle gets enlarged. 
This demonstrates that the $L_2$BN can make the intra-class features more compact and the inter-class features more separable. 
Therefore, the $L_2$BN can enhance the discriminative ability of neural networks, which is the reason for the accuracy improvement.  

\begin{figure}[t]
  \centering
  \includegraphics[width=\columnwidth]{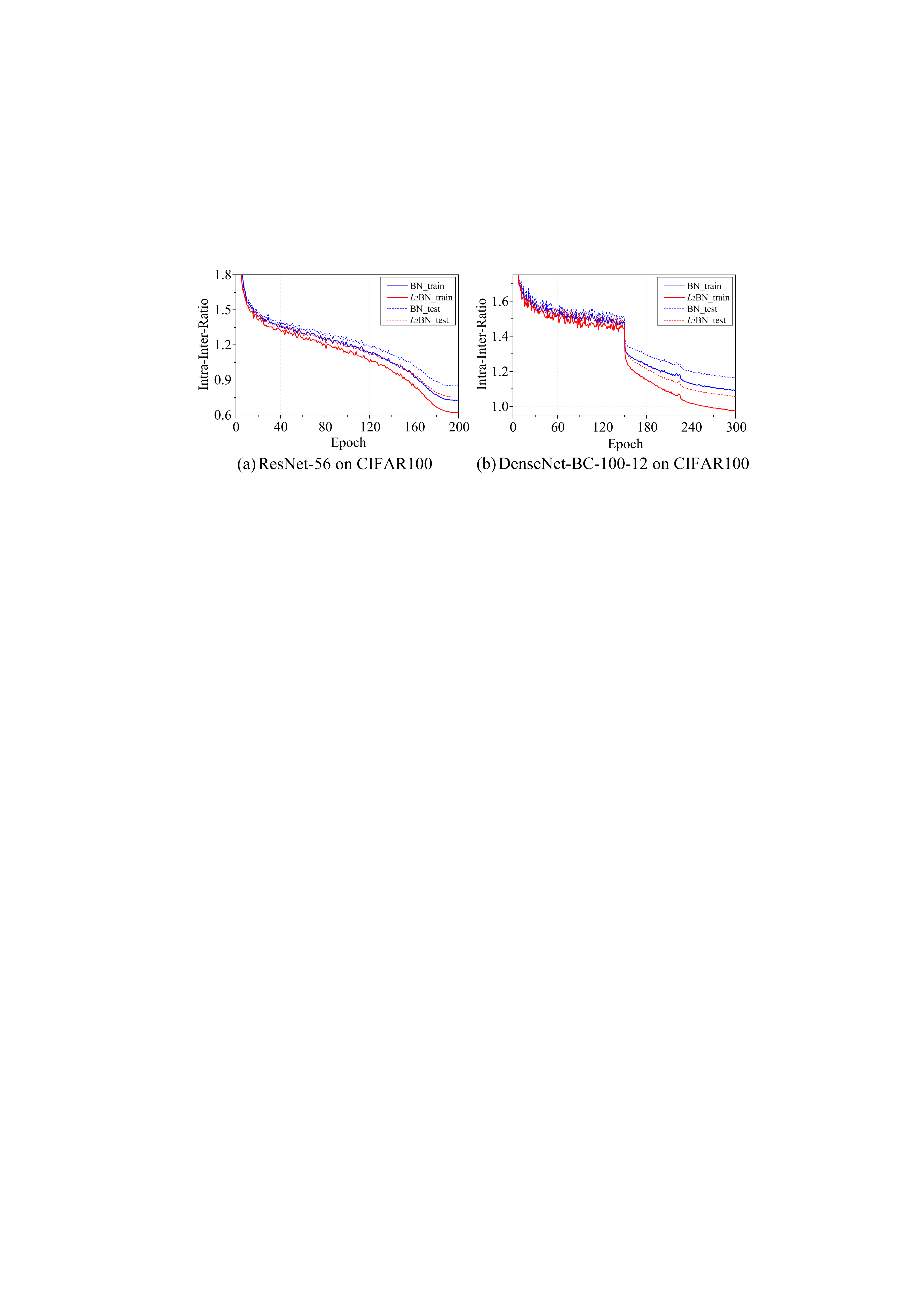}
  \caption{The comparison of $IIR$ curves. On both training data and test data, $L_2$BN achieves consistently smaller $IIR$.}
  \label{fig:IIR_curves}
  \vspace*{-3mm}
\end{figure}

Given the relative relationship between intra-class compactness and inter-class discrepancy, using either intra-angle or inter-angle alone is not sufficient to evaluate the discriminability of a model. 
For that reason, we define the Intra-angle and Inter-angle Ratio, abbreviated as $IIR$, as follows:
\begin{equation}\label{IIR}
  IIR = \frac{intra\mbox{-}angle}{inter\mbox{-}angle} .
\end{equation}
The $IIR$ is a unified metric to evaluate the intra-class compactness and the inter-class discrepancy, the smaller the better.  
We use the unified $IIR$ to verify the effect of $L_2$BN.  
For all the backbones, the $L_2$BN can always achieve significantly smaller $IIR$ on both training data and test data, as shown in Table~\ref{tab:results_cifar100}. 
To be more rigorous, we compare the $IIR$ of BN and $L_2$BN throughout the training process in Figure~\ref{fig:IIR_curves}. 
We can see that the $L_2$BN can continuously get smaller $IIR$, demonstrating the advantages of $L_2$BN. 

It is worth mentioning that the types of backbones used here are diverse, including models with skip connection like ResNet~\cite{he2016deep} and  DenseNet~\cite{huang2017densely}, models without skip connection like VGG~\cite{simonyan2014very}, and models searched by Neural Architecture Search like RegNet~\cite{radosavovic2020designing}. 
This indicates that, as a simple plug-and-play module, the $L_2$BN is shown to be
architecture-agnostic and produce persistent performance improvement. 
Therefore, the key advantage of $L_2$BN highlighted in this paper is not the significance but
the robustness and generalizability.

\noindent\textbf{Choice of Feature Normalization.}
In the Method Section~\ref{method}, we employ the $l_2$ normalization on the whole feature map of each sample by default.  
To justify our choice, we also evaluate in Table~\ref{tab:l2_choice} other possible alternatives to $l_2$ normalization such as Layer Normalization~\cite{ba2016layer}, Instance Normalization~\cite{ulyanov2016instance}, and Position Normalization~\cite{li2019positional}, denoted by LNBN, INBN, and PNBN, respectively.


\begin{table}[t]
  \centering
  \setlength\tabcolsep{9.0pt}
  \begin{tabular}{lccc}
    \toprule
     \multirow{2}{*}{Methods}  
     &\multirow{2}{*}{Accuracy(\%)$\uparrow$}
     &\multicolumn{2}{c}{$IIR$}   \\
     \cmidrule(lr){3-4}
     &&train$\downarrow$ & test$\downarrow$
    \\
    \midrule
    BN(baseline)  & 72.88$\pm$0.25  & 0.726  & 0.848  \\ 
    LNBN  & 73.57$\pm$0.20  & 0.674  & 0.802  \\ 
    INBN  & 69.34$\pm$0.29  & 0.739  & 0.870  \\ 
    PNBN  & 72.49$\pm$0.17  & 0.714  & 0.846  \\ 
    $L_2$BN(ours)  & \textbf{73.92}$\pm$0.20  & \textbf{0.638}  & \textbf{0.772}  \\
    \bottomrule
  \end{tabular}
  \caption{Ablation of possible alternatives to $l_2$ normalization in $L_2$BN using ResNet-56 on CIFAR100.}
  \label{tab:l2_choice}
  \vspace*{-3mm}
\end{table}

We observe that the INBN and PNBN get much poorer performance than $L_2$BN, in terms of both accuracy and $IIR$. 
We argue that it is the entire feature map that represents the sample feature, whether the Instance Normalization or the Position Normalization destroys the sample-level feature information. 
The Layer Normalization is also performed on the whole feature map. 
Compared with the $l_2$ normalization, the main difference is that Layer Normalization has a centering operation. 
From the analysis in Figure~\ref{fig:inter_l2bn} and Figure~\ref{fig:intra_l2bn}, the centering operation is unnecessary for the intra-class compactness and inter-class discrepancy. 
Moreover, the centering operation changes the orientation of feature vector and thus may change what the feature vector can represent~\cite{ioffe2015batch}. 
To address this, Layer Normalization in the original paper~\cite{ba2016layer} employs a learned affine transformation to adaptively compensate for the possible information loss. 
However, the subsequent Batch Normalization makes the learned affine transformation of Layer Normalization ineffective in LNBN. 
Therefore, the information loss caused by Layer Normalization can not be effectively compensated. 
The $l_2$ normalization does not change the orientation of feature vector and therefore does not related to the semantic information loss~\cite{liu2017deep, liu2018decoupled}. 
The results in Table~\ref{tab:l2_choice} also prove that  both the accuracy and the $IIR$ of LNBN are inferior to those of $L_2$BN.  
In summary, compared with these possible alternatives, $l_2$ normalization is the best choice.

\noindent\textbf{Applying $L_2$BN to Different  Layers.} 
As described in Section~\ref{implementation}, the $L_2$BN is applicable to all layers. 
To confirm this point, we study the effect of $L_2$BN applied to different parts of neural networks, as shown in Table~\ref{tab:L2BNtolayers}. 
Even if the $L_2$BN is only used for part of the neural network, the accuracy can be improved to varying degrees. 
However, the accuracy is enhanced to the greatest extent when applying $L_2$BN to all layers. 
This indicates that the $L_2$BN is effective for both
the output layer and the hidden layers, and the effects can be
accumulated.

\begin{table}[t]
  \centering
  \setlength\tabcolsep{9pt}
  \begin{threeparttable}
  \begin{tabular}{lc}
    \toprule
    \multicolumn{1}{c}{$L_2$BN}       &Accuracy(\%)  \cr
    \midrule
    None (baseline)       & 72.88$\pm$0.25    \\
    Only classification layer       & 73.04$\pm$0.27    \\
    ResStage 1                      & 73.59$\pm$0.27    \\
    ResStage 2-3                    & 73.43$\pm$0.17    \\
    All layers                  & \textbf{73.92}$\pm$0.20        \\
    \bottomrule
  \end{tabular}
  \end{threeparttable}
  \caption{Ablation of applying $L_2$BN to different layers using ResNet-56 on CIFAR100.}
  \label{tab:L2BNtolayers}
  \vspace*{-3mm}
\end{table}

\noindent\textbf{Comparison with Other Normalization Methods.}
This section compares $L_2$BN with several existing variants of Batch Normalization, including Layer Normalization (LN)~\cite{ba2016layer}, Instance Normalization (IN)~\cite{ulyanov2016instance}, Group Normalization (GN)~\cite{wu2018group}, Switchable Normalization (SN)~\cite{luo2018differentiable}, Batch-Channel Normalization (BCN)~\cite{qiao2019micro}, and Batch-Instance Normalization (BIN)~\cite{nam2018batch}. 
For GN, we test settings with the channel-per-group (cpg) values of 4 and 8. 
For BCN, we set the number of groups to min\{32, (the number of channels)/4\}, as in the original paper~\cite{qiao2019micro}. 
The results are shown in Table~\ref{tab:normalizations}. 
For LN, IN, and GN, they abandon the batch statistics and thus perform worse than BN in popular training settings~\cite{wu2018group}. 
SN, BCN, and BIN can be used to address the problem of poor performance of BN when the mini-batch size is very small. 
However, their performance in popular training settings is on par with or worse than BN. 
In contrast, the $L_2$BN can still enhance BN when the mini-batch size is not small. 
Overall, the $L_2$BN performs better than these normalization methods without any hyper-parameters.

\begin{table}[h]
  \vspace*{-2mm}
  \centering
  \setlength\tabcolsep{7pt}
  \begin{tabular}{lccc}
    \toprule
     \multirow{2}{*}{Methods}  
     &\multirow{2}{*}{Accuracy(\%)$\uparrow$}
     &\multicolumn{2}{c}{$IIR$}   \\
     \cmidrule(lr){3-4}
     &&train$\downarrow$ & test$\downarrow$
    \\
    \midrule
    BN(baseline)~\cite{ioffe2015batch}  & 72.88$\pm$0.25  & 0.726  & 0.848  \\ 
    LN~\cite{ba2016layer}  & 69.88$\pm$0.15  & 0.877  & 0.982  \\ 
    IN~\cite{ulyanov2016instance}  & 69.77$\pm$0.20  & 0.737  & 0.871  \\ 
    GN-cpg4~\cite{wu2018group}  & 70.35$\pm$0.19  & 0.746  & 0.881  \\ 
    GN-cpg8~\cite{wu2018group}  & 70.68$\pm$0.51  & 0.773  & 0.895  \\ 
    SN~\cite{luo2018differentiable}  & 72.85$\pm$0.17  & 0.691  & 0.816  \\ 
    BCN~\cite{qiao2019micro}  & 72.90$\pm$0.20  & 0.795  & 0.909  \\ 
    BIN~\cite{nam2018batch}  & 70.83$\pm$0.14  & 0.809  & 0.937  \\ 
    $L_2$BN(ours)  & \textbf{73.92}$\pm$0.20  & \textbf{0.638}  & \textbf{0.772}  \\
    \bottomrule
  \end{tabular}
  \caption{Comparison with other normalization methods using ResNet-56 on CIFAR100.}
  \label{tab:normalizations}
  \vspace*{-2mm}
\end{table}

\vspace*{-2mm}
\subsection{Acoustic Scene Classification}
\vspace*{-1mm}

\noindent\textbf{Experimental Settings.}
To further verify the effectiveness of our proposed method, we conduct experiments on the acoustic scene classification task.
We experiment on the TUT Urban Acoustic Scenes 2020 Mobile Development dataset~\cite{toni2020tau}, which consists of 10-seconds audio segments from 10 acoustic scenes and contains in total of 64 hours of audio. 
The task we choose is a subtask of the acoustic scene classification in the challenge on detection and classification of acoustic scenes and events~(DCASE)~\cite{DCASE2020}.
The goal is to classify the audio into $10$ distinct specific acoustic scenes, including airport, public square and urban park, etc.

We employ three CNN-based architectures as backbone models, including ResNet-17~\cite{MD2020Acoustic}, FCNN~\cite{hu2020devicerobust}, and fsFCNN~\cite{hu2020devicerobust}. 
The optimizer is SGD with a cosine-decay-restart learning rates scheduler, in which the maximum and minimum learning rates are 0.1 and 1e-5 respectively. 
We train the ResNet-17 for 126 epochs and the FCNN and fsFCNN for 255 epochs. 
All of them are trained with a batch size of 32.
For a fair comparison, we train both the $L_2$BN models and the corresponding BN models from scratch under the same configurations.

\begin{table}[t]
    \centering
    \setlength\tabcolsep{9pt}
    \begin{tabular}{llc}
    \toprule
    \multicolumn{1}{c}{Model}  & BN/$L_2$BN  & Accuracy(\%)    \\
    \midrule
    ResNet-17  & BN  & 72.51  \\
    ResNet-17  & $L_2$BN  & \textbf{72.64}    \\
    \hline
    FCNN  & BN  & 69.17  \\
    FCNN  & $L_2$BN  & \textbf{70.55}  \\
    \hline
    fsFCNN  & BN  & 71.19    \\
    fsFCNN  & $L_2$BN  & \textbf{72.44}   \\
    \bottomrule
    \end{tabular}
    \caption{Results of Acoustic Scene Classification.}
    \label{tab:result_acoustic}
    \vspace*{-4mm}
\end{table}

\noindent\textbf{Results.} 
Table~\ref{tab:result_acoustic} shows the comparison results of the baseline models and their $L_2$BN models on the acoustic scene classification task. 
For all three baseline models, the corresponding $L_2$BN models can boost the accuracy under the same configurations. 
Specifically, $L_2$BN can achieve a significant accuracy improvement of more than 1\% for FCNN and fsFCNN. 
Together with the results of image classification experiments, we conclude that the $L_2$BN is effective for different domains, which indicates that our proposed method is scalable and general.

\section{Conclusion} 
In this paper, we propose a strong substitute for batch normalization, 
the $L_2$BN, which makes the $l_2$ norms of sample features identical before feeding them into BN. 
Our analysis and experiments reveal that the proposed $L_2$BN can facilitate intra-class compactness and inter-class discrepancy. 
Besides, the characteristic of requiring no additional parameters and hyper-parameters makes it easy to use. 
We evaluate the effect of $L_2$BN on image classification and acoustic scene classification tasks with various deep neural networks, demonstrating its effectiveness and generalizability. 
As a simple but effective operation, we believe that $L_2$BN can be integrated into a wide range of application scenarios as a plug-and-play module without any tuning.


{\small
\bibliographystyle{ieee_fullname}
\bibliography{egbib}
}

\includepdfmerge{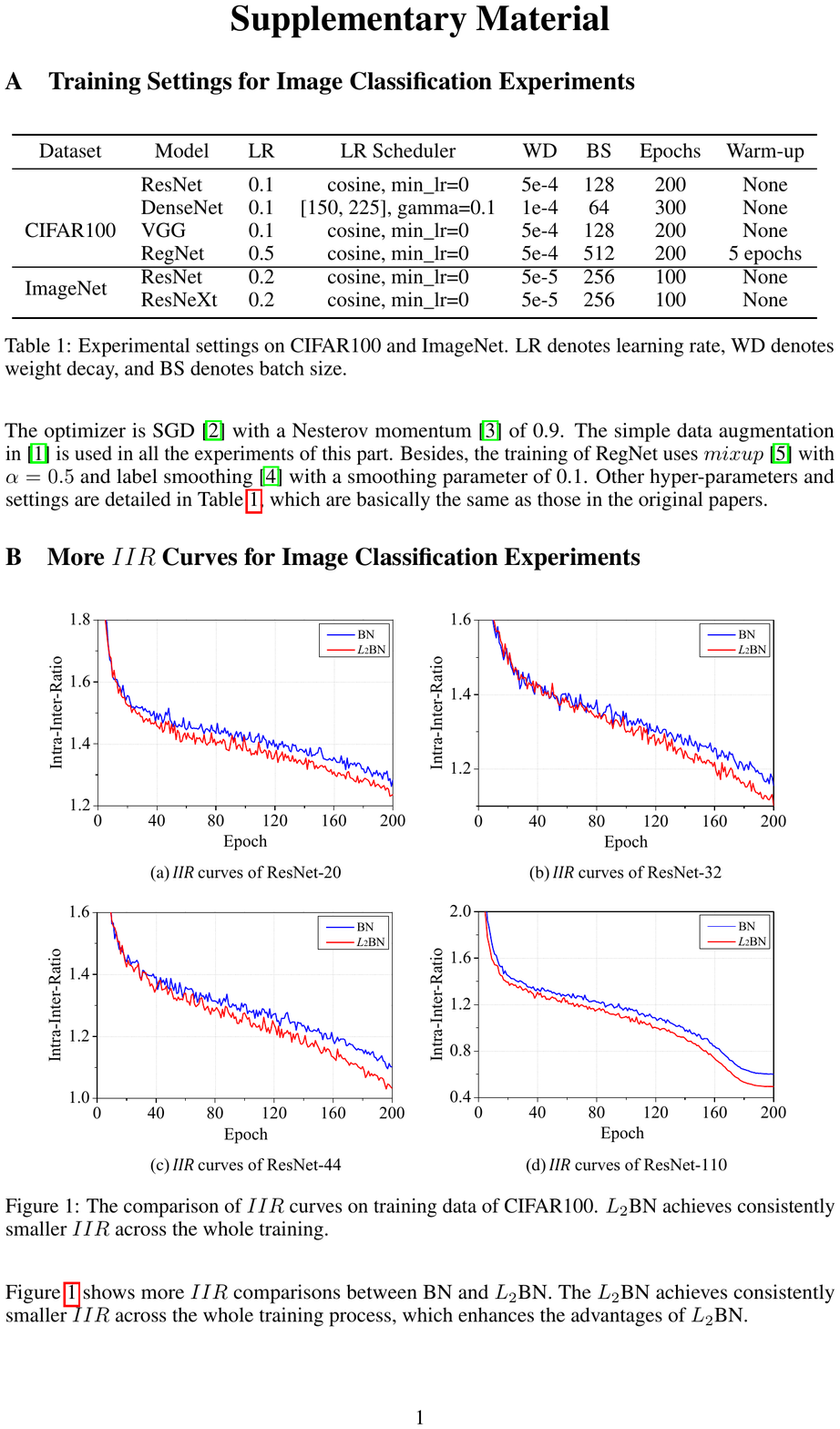,1}
\includepdfmerge{materials/SupplementaryMaterial.pdf,2}

\end{document}